\pdfoutput=1

\documentclass[11pt]{article}

\usepackage[]{EMNLP2023}

\usepackage{times}
\usepackage{enumitem}
\usepackage{booktabs}
\usepackage{latexsym}
\usepackage{tikz}
    \usetikzlibrary{shadows}
\usepackage{tcolorbox}
    \tcbuselibrary{skins}
\definecolor{black}{rgb}{0, 0, 0}

\usepackage{booktabs}
\usepackage{multirow}

\usepackage{algorithm}
\usepackage{algpseudocode}
\usepackage{booktabs}

\usepackage{graphicx}
\usepackage{caption}
\usepackage{subcaption}
\usepackage{lipsum}
\usepackage{float}
\usepackage{setspace}
\usepackage{array}

\newcommand{\mypar}[1]{\vspace{2mm}\noindent\textbf{#1. }}
\newcommand{\fone}{\texttt{F1-score}}
\newcommand{\preci}{\texttt{Precision}}
\newcommand{\rec}{\texttt{Recall}}

\newcommand{\semihalf}{\texttt{semi-half}}
\newcommand{\permutation}{\texttt{permutation}}
\newcommand{\permutationr}{\texttt{permutation-R}}
\newcommand{\permutationq}{\texttt{permutation-Q}}
\newcommand{\ngram}{\texttt{n-gram}}

\usepackage{geometry}
\newtcolorbox{mybox}[1]{
    enhanced,
    borderline={2pt}{0mm}{black},
    borderline={.7pt}{1mm}{black},
    fonttitle=\bfseries,
    title=#1,
    arc=3mm,
    segmentation hidden,
    top=8pt,
    bottom=8pt
}

\usepackage{xcolor}
\definecolor{Apricot}{rgb}{0.98, 0.81, 0.69}

\newcommand{\highLight}[2][Apricot]{%
  \begingroup
  \setlength{\fboxsep}{1pt}%
  \colorbox{#1}{#2}%
  \endgroup
}

\usepackage[T1]{fontenc}
\usepackage{amsmath} 
\usepackage{hyperref}

\usepackage[utf8]{inputenc}

\usepackage{microtype}

\newcommand{\DD}{\phantom{0}}

\usepackage{inconsolata}

\title{Simulating Training Data Leakage in Multiple-Choice Benchmarks \\for LLM Evaluation}

\author{Naila Shafirni Hidayat$^{1}$ \quad Muhammad Dehan Al Kautsar$^{2}$ \\ \textbf{Alfan Farizki Wicaksono}$^{1}$  \quad \textbf{Fajri Koto}$^{2}$ \\ 
        $^{1}$Faculty of Computer Science, Universitas Indonesia \\
$^{2}$Department of Natural Language Processing, MBZUAI \\
	\texttt{\small naila.shafirni@ui.ac.id, alfan@cs.ui.ac.id,
    \{muhammad.dehan,fajri.koto\}@mbzuai.ac.ae 
    } 
}

\begin{document}

\maketitle
\begin{abstract}

The performance of large language models (LLMs) continues to improve, as reflected in rising scores on standard benchmarks. However, the lack of transparency around training data raises concerns about potential overlap with evaluation sets and the fairness of reported results. Although prior work has proposed methods for detecting data leakage, these approaches primarily focus on identifying outliers and have not been evaluated under controlled simulated leakage conditions. In this work, we compare existing leakage detection techniques, namely {\permutation} and {\ngram}-based methods, under a continual pretraining setup that simulates real-world leakage scenarios, and additionally explore a lightweight method we call {\semihalf} question. Although {\semihalf} offers a low-cost alternative, our analysis shows that the n-gram method consistently achieves the highest {\fone}. We also refine these techniques to support instance-level detection and reduce computational overhead. Leveraging the best-performing method, we create cleaned versions of MMLU and HellaSwag, and re-evaluate several LLMs. Our findings present a practical path toward more reliable and transparent evaluations, and we recommend contamination checks as a standard step before releasing benchmark results.\footnote{Code and dataset are available at \url{https://github.com/nailashfrni/mcq-leakage-detection-code}}

\end{abstract}

\section{Introduction}

The development of Large Language Models (LLMs) has shown competitive performance 
on 
multiple-choice question answering \cite{brown2020languagemodelsfewshotlearners, openai2024gpt4technicalreport, qwen2025qwen25technicalreport, gemmateam2024gemmaopenmodelsbased, grattafiori2024llama3herdmodels}.
These models are evaluated on benchmark datasets designed to assess specific competencies such as knowledge and reasoning.
However, many LLMs do not disclose their pre-training data \citep{piktus-etal-2023-roots}, raising concerns that benchmark evaluation sets were included in training.

This lack of transparency raises a critical question: \textit{Do current evaluation results reflect the true generalization abilities, or are they barely a product of memorization?} 
Suppose a model has been trained on evaluation datasets during training. In that case, it doubts the fairness of comparisons as its ability to answer questions might originate from data memorization \cite{carlini2023quantifyingmemorizationneurallanguage} rather than reasoning or generalization. 
This issue, referred to as data contamination \cite{magar-schwartz-2022-data, balloccu-etal-2024-leak} poses significant concerns for reliable benchmarking. 

Recent studies have introduced various methods to detect data contamination in multiple-choice question (MCQ) benchmarks for LLMs \cite{ni2024training, xu2024benchmarking, li2023estimatingcontaminationperplexityquantifying}. However, these approaches primarily focus on identifying outliers and have not been systematically evaluated under controlled, simulated leakage conditions. Furthermore, there is limited understanding of their relative effectiveness \cite{hu2022membershipinferenceattacksmachine, samuel2024datacontaminationdetectionmodern, fu2025doesdatacontaminationdetection}, and no consensus on the optimal configurations for detecting training data leakage. To address these gaps, we benchmark existing leakage detection methods under controlled simulations, focusing on two widely used MCQ datasets in LLM evaluation: the Massive Multitask Language Understanding (MMLU) dataset \citep{hendryckstest2021} and the HellaSwag dataset \citep{zellers2019hellaswagmachinereallyfinish}.

Our key contributions are as follows:
\begin{itemize}
    \item We compare three leakage detection methods under simulated training data leakage via continual pre-training: (1) the {\semihalf} method, which tests whether a truncated version of a question still results in the correct answer; (2) the {\permutation} method, originally proposed by \citet{ni2024training}, which evaluates whether the original option order yields the highest likelihood among all permutations; and (3) the {\ngram} method, which assesses the similarity between a generated option sentence and the original, following \citet{xu2024benchmarking}.   
    \item We improve the {\permutation} method by introducing two variants, {\permutationr} and {\permutationq}, which reduce computational overhead while improving {\fone}. We also refine the {\ngram} method to support instance-level detection.
    \item We construct and release a subset of the MMLU and HellaSwag dataset verified to be free of contamination across several popular LLMs. Furthermore, we re-evaluate these models on the clean subset to observe shifts in performance ranking. 
\end{itemize}
\section{Related Work}

\subsection{Evaluation Benchmark of LLM}
Language model evaluation has shifted from classical NLP tasks—such as named entity recognition and part-of-speech tagging—toward benchmarks that assess knowledge and reasoning, driven by advances in fluency and coherence. These evaluations commonly adopt a multiple-choice format, exemplified by MMLU \citep{hendryckstest2021}, which compiles questions of $57$ subjects from a wide range of school exams across different education levels. Other popular reasoning benchmarks include HellaSwag \citep{zellers2019hellaswagmachinereallyfinish}, PIQA \citep{Bisk2020}, BoolQ \citep{clark-etal-2019-boolq}, Social--IQa (SIQA) \citep{sap2019socialiqacommonsensereasoningsocial}, and TruthfulQA \citep{Lin2021TruthfulQAMH}.

MMLU is one of the most widely used datasets for evaluating the knowledge capabilities of LLMs. To improve its quality and robustness, prior work has introduced several variants. MMLU-Pro \citep{wang2024mmluprorobustchallengingmultitask} enhances the dataset by increasing question difficulty through filtering, expanding answer choices from four to ten, and incorporating expert review. Separately, \citet{gema2025mmlu} released MMLU-Redux, a cleaned version that addresses issues such as ambiguous phrasing, multiple correct answers, and incorrect ground truths. However, despite these improvements, both variants primarily focus on question quality and coverage. Neither MMLU-Pro nor MMLU-Redux incorporates systematic filtering or analysis to detect overlap with pretraining data, leaving open the risk that benchmark scores may reflect memorization rather than true generalization.

\subsection{Data Contamination Detection}
Numerous methods have been proposed to detect data contamination in LLMs, broadly falling into logit-based, generation-based, and hybrid categories. Logit-based methods analyze the model’s output probabilities or internal states; for example, \citet{ni2024training} compare log-probabilities across different option orders, while \citet{li2023estimatingcontaminationperplexityquantifying} use perplexity to detect dataset-level leakage. However, these approaches primarily focus on outlier detection, offer limited support for instance-level analysis, and have not been evaluated under controlled training leakage simulations. In contrast, generation-based methods assess whether the model can reconstruct reference content when prompted. \citet{golchin2024timetravelllmstracing} use ``time-travel'' prompts incorporating dataset-specific cues to regenerate partial instances and compare them to the original text. \citet{xu2024benchmarking} introduce a hybrid approach combining n-gram similarity with perplexity, though their focus is on GSM8K \cite{cobbe2021gsm8k} and MATH datasets \cite{hendrycksmath2021}. Importantly, these methods have not been tested on the multiple-choice question (MCQ) format, which remains the most widely used prominent structure in LLM evaluation benchmarks.
\section{Leakage Detection Method}

\begin{figure*}[ht]
  \centering
  \includegraphics[width=\linewidth]{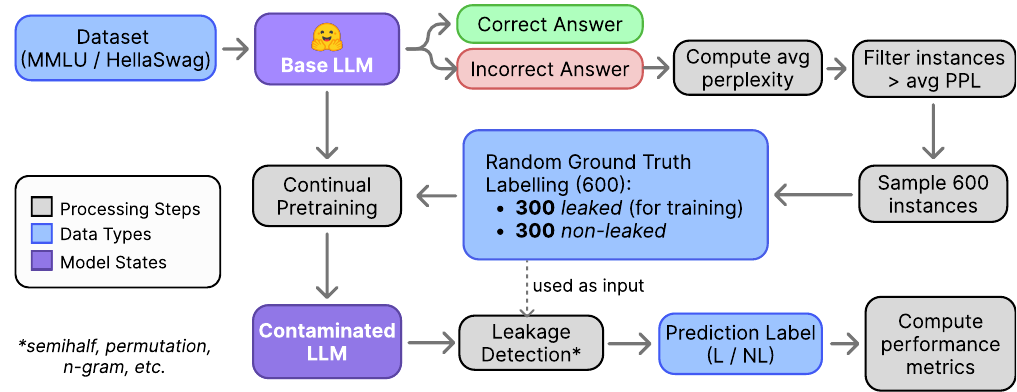}
  \caption{Workflow for simulating data leakage and evaluating detection methods. Boxes represent different components: processing steps (gray), data types or datasets (blue), and model states (purple).}
  \label{fig:experiment-design}
\end{figure*}

We simulate training data leakage using continual pre-training and compare the effectiveness of three detection methods: {\semihalf}, {\permutation} \citep{ni2024training}, and {\ngram} \citep{xu2024benchmarking}. To improve efficiency, we introduce a simplified variant of {\permutation}, called {\permutationr}, and propose a new method, {\permutationq}, built on the same foundation.

\subsection{Leakage Simulation}

Figure~\ref{fig:experiment-design} illustrates our controlled simulation of intentional data leakage. We start by selecting questions from MMLU~\cite{hendrycks2021measuringmassivemultitasklanguage} and HellaSwag~\cite{zellers2019hellaswagmachinereallyfinish} that the model initially answers incorrectly. From this set, we randomly sample $600$ instances with above-average perplexity to ensure unfamiliarity and minimize the chance of prior exposure during pre-training. We use $300$ of these samples for continual pre-training via Low-Rank Adaptation (LoRA)~\cite{hu2021loralowrankadaptationlarge}, simulating data leakage. After training, all detection methods are applied to the full set of $600$ instances. The $300$ examples included in pre-training are labeled as ``Leaked'', while the remaining $300$ serve as ``Not Leaked''. We assess detection performance using {\preci}, {\rec}, and {\fone}. 

\subsection{\texttt{Semi-half} Detection Method}

\begin{figure}[t]
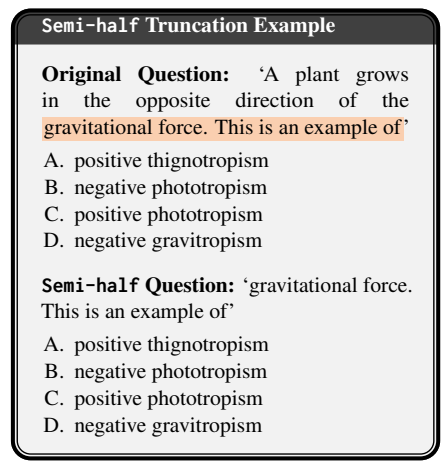

    \centering
    \resizebox{0.75\linewidth}{!}{%
        \begin{mybox}{\texttt{Semi-half} Truncation Example}
        \textbf{Original Question:} `A plant grows in the opposite direction of the  \highLight[Apricot]{gravitational force. This is an example of}'
        \begin{itemize}[label={}, leftmargin=1.5em, itemsep=0em, topsep=0.3em, parsep=0em, partopsep=0em]
            \item[A.] positive thignotropism
            \item[B.] negative phototropism
            \item[C.] positive phototropism
            \item[D.] negative gravitropism
        \end{itemize}
    
        \vspace{2mm}
        \textbf{\texttt{Semi-half} Question:} `gravitational force. This is an example of'
        \vspace{0.3em}
        \begin{itemize}[label={}, leftmargin=1.5em, itemsep=0em, topsep=0em, parsep=0em, partopsep=0em]
            \item[A.] positive thignotropism
            \item[B.] negative phototropism
            \item[C.] positive phototropism
            \item[D.] negative gravitropism
        \end{itemize}
        \end{mybox}
    }
    \caption{Semi-half truncation example\label{fig:semihalf-example}}
    \vspace{-0.5cm}
\end{figure}

To answer a multiple-choice question, a model relies on both the question and the options~\citep{robinson2023leveraginglargelanguagemodels}. If it can still select the correct answer after the first half of the question is removed, this may suggest prior exposure during pre-training. Motivated by this, we propose a simple truncation-based method that retains only the final seven words of each question, providing minimal context while aligning with the autoregressive nature of decoder-based LLMs. The seven-word limit reflects the average half-length of the MMLU questions. Figure~\ref{fig:semihalf-example} illustrates a semi-half truncation example: if the model has seen the question during pre-training, it may still produce the correct answer despite the limited input; otherwise, the model is unlikely to produce the correct answer due to insufficient context.

\subsection{{\texttt{Permutation} Method}}

\citet{ni2024training} proposed a method to detect contamination by evaluating how a language model assigns probabilities across different orders of multiple-choice answer options. The key idea is that if the model consistently assigns the highest probability to the original option order (e.g., A-B-C-D), it may have memorized that specific multiple-choice instance during training and indicated potential contamination. 

The detailed method is explained in Appendix~\ref{sec:detailed-permutation}. The algorithm complexity for this method is dominated by computing log-probability scores for each option order variation in a question. In big-Oh notation, the complexity is stated as
$O(n!)$,
where $n$ denotes the number of options. This is considered a costly approach, and we modified this method to better achieve a less complex algorithm.

\mypar{\texttt{Permutation-R}}
Our main concern with the {\permutation} method is its computational cost in computing the log-probability for all permutation variations. To address this, we eliminate permutations that have nearly similar log-probability distributions for all questions, then retain only a representative permutation subset.

To determine which permutation pairs show similar distributions, we employ Mean Absolute Difference (MAD) to measure the discrepancy in log-probability scores between two permutations. Let $p_{ji}$ and $p 
_{ki}$ represent the log-probability scores for permutations $j$ and $k$ on question $i$, respectively, and let $z$ denote the number of questions. The mean absolute difference between permutations $j$ and $k$, denoted by $\texttt{Diff}(j,k)$ is computed as:
\[\texttt{Diff}(j,k) = \frac{1}{z} \sum_{i=1}^{z} |p_{ji} - p_{ki}| \, .\]

We experiment with three different models: Qwen2.5-7B \citep{qwen2025qwen25technicalreport}, LLaMA-3.1-8B \citep{touvron2023llamaopenefficientfoundation}, and Gemma-7B \citep{gemma2}. For each experiment, we compute $\texttt{Diff}(j, k)$ for all possible permutation pair $j$ and $k$ and average the ranking across experiment. Since lower MAD indicates more similar distribution, we sort the average rank in increasing order. From that order, we retain only one permutation from each pair. To determine the optimal number of permutations used, we experiment with various proportion values $p$ to observe which setting best balances computational cost and performance. The optimal $p$ is then selected and used as the final configuration for the {\permutationr} method.

The algorithm complexity is $O\left(p.[n!]\right)$, where $n$ denotes the number of options and $p$ is for percentage of permutations used. This improvement might not be significant in the big-Oh notation since it still has permutation complexity. However, in practice, the reduced variation factor contributes to reducing computation time. 

\mypar{\texttt{Permutation-Q}}

In practice, {\permutationr} improves efficiency by introducing a fractional term upfront. However, challenges arise when dealing with tasks that involve more than four answer choices, such as MMLU-Pro \cite{wang2024mmluprorobustchallengingmultitask}, with $10$ options. To address this, we propose {\permutationq} method, that replaces the factorial component with a more tractable quadratic approximation. The idea is to employ only two options in each log-probability calculation. 

Suppose that we have an instance $x=[q, o_1, o_2, ..., o_n]$ where $q$ denotes the question and $o_n$ is the last option answer. We generate permutation $P^n_2$ from $o=\{o_1, o_2, ..., o_n\}$ to only two options. We calculate the log-probability score for all possible permutations of two options. If the original option order (A-B) produces the maximum log-probability among all orders, we consider the instance $x$ as `Leakage', otherwise not.\footnote{The key idea is to compare the original 2-option pairs with its permutations, regardless of whether the correct answer is present in the pair.} The algorithm is presented in Algorithm \ref{alg:permutation-Q} in Appendix~\ref{sec:permutation-q-algorithm}.

The complexity of the above method is centered on log-probability calculation for all possible combinations of options. The big-Oh notation is computed as:
\[ O(P^n_2) = O(n. (n-1)) = O(n^2-n)=O(n^2).\]
The complexity is reduced from factorial to quadratic, which is an improvement in detecting a leakage in a particular model.

\subsection{\texttt{N-Gram} Method}

The {\ngram} method builds on the approach introduced by \citet{xu2024benchmarking}, which uses n-gram accuracy to detect potential data contamination during pre-training. The core idea is to test whether a model has memorized benchmark answer options by evaluating its ability to generate them. While the original method generates $n$ tokens per prompt and compares them to a reference sequence, we modify it to generate an entire option sentence in a single inference. Other than that, while \citet{xu2024benchmarking} focuses on detecting leakage at the dataset-split level by comparing metric differences between original and synthetic data, we adapt it to work at the instance level. This modification allows the method to identify contamination on a per-example basis and allows the analysis to be more comprehensive. The full details of the method and its algorithm are presented in Appendix~\ref{sec:ngram-algorithm}.
\section{Experiment}

\subsection{Set-Up}
We experiment with four models and two evaluation benchmarks—MMLU~\cite{hendrycks2021measuringmassivemultitasklanguage} and HellaSwag~\cite{zellers2019hellaswagmachinereallyfinish}—to simulate data contamination in LLMs. The model list detailed in Table~\ref{tab:model-sources} in Appendix~\ref{sec:model-setting}. Each model undergoes continual pre-training on each benchmark. Using the Adam optimizer, we set the learning rate to $1e-5$ for the language model head and $5e-4$ for other parameters. Each model is trained for $10$ epochs with a weight decay of $0.01$, a warmup ratio of $0.1$, and a cosine learning rate scheduler. We also record the loss at each epoch for monitoring. For experiments involving LLaMA-3.1-8B base and Gemma-7B, we utilize an H100 SXM GPU with 80GB VRAM. For all other models, we use an A40 GPU with 48GB VRAM.

After completing the continual pre-training for all eight settings, we tune the threshold of {\ngram} and optimize the {\permutation} method first. We then use this configuration to evaluate all methods and compare their performance.

\subsection{Preliminary Results}
\begin{figure}[t]
  \centering
  \includegraphics[width=\linewidth]{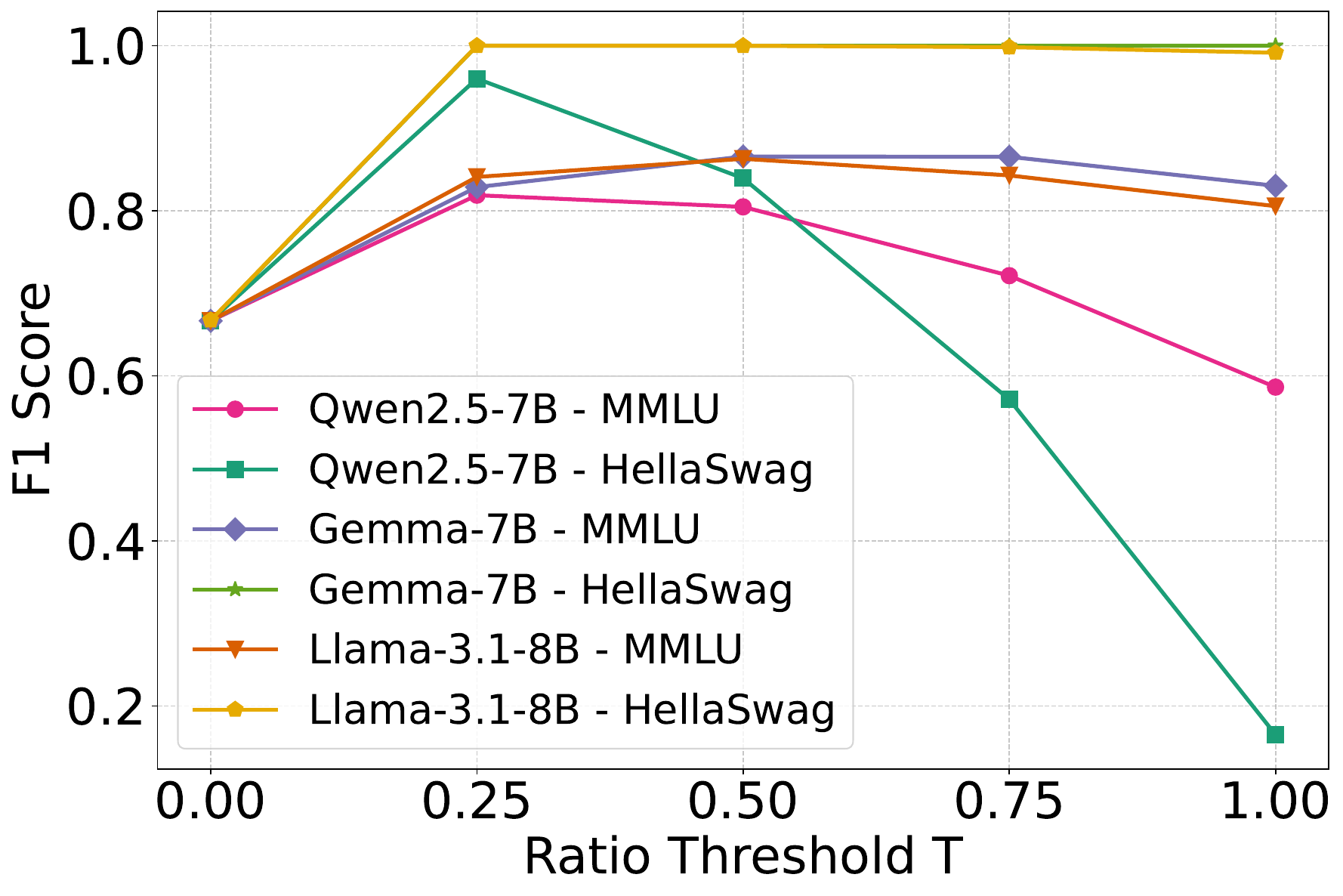}
  \caption{Comparison of {\ngram} detection F1-score performance under varying ratio thresholds $T$.}
  \label{fig:ngram-threshold}
\end{figure}

\mypar{Varying \texttt{N-Gram} Method's Threshold}
We explore the effect of varying the threshold $T$ in the {\ngram} method, which determines its sensitivity to determine an instance as `Leakage'. The results of this comparison are presented in Figure~\ref{fig:ngram-threshold}. Across all experiments, a threshold of $T=0.25$ consistently yields the best or comparable {\fone}. Notably, in Qwen2.5-7B on HellaSwag, {\fone} drops sharply as $T$ increases. For LLaMA-8B and Gemma-7B on HellaSwag, {\fone} remains at $100\%$ for $T=0.25$, $0.5$, and $0.75$, with only a slight decrease at $T=1.0$. A similar trend appears in MMLU, where {\fone} peaks at $T=0.5$ but still exceeds $80\%$ at $T=0.25$.

These results suggest that $T=0.25$ offers the best balance of sensitivity and reliability for detecting data contamination. This approach ensures that all potentially suspicious questions, even if only a single option is successfully replicated, are treated as `Leakage'. This allows us to capture as many contaminated instances as possible.

\mypar{Reducing Permutation Variation}
Using the Mean Absolute Difference (MAD), we compute the difference scores between log-probability distributions for each permutation pair. We rank all pairs by similarity for each model and average these rankings to identify the top 24 most similar pairs (see Table~\ref{tab:rank-diff} in Appendix~\ref{sec:mad-table}). Notably, many of these differ by only two character swaps, suggesting such changes have minimal effect on the log-probabilities.

\begin{figure}[t]
  \centering
  \includegraphics[width=\linewidth]{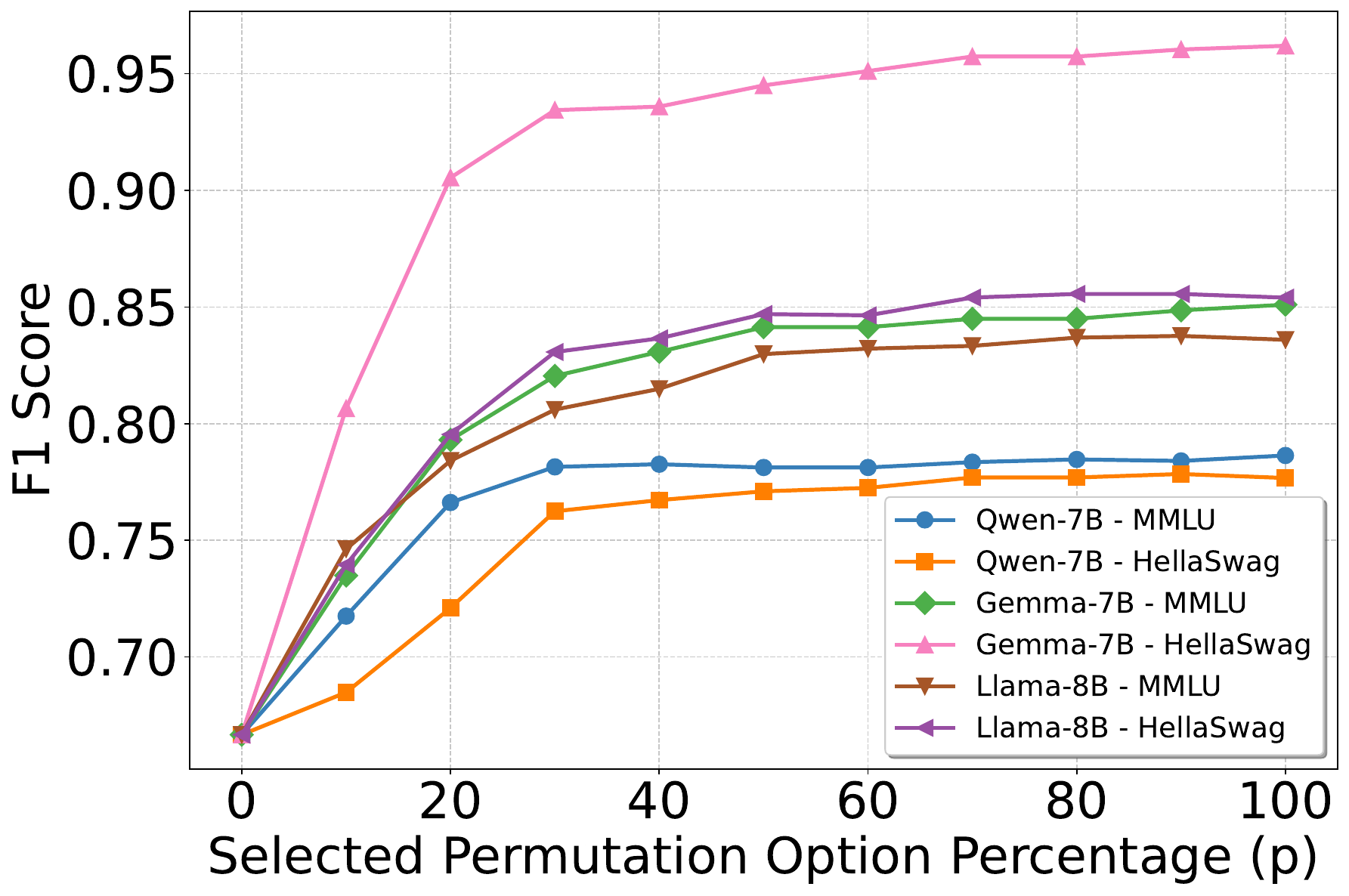}
  \caption{Performance of {\permutation} at different percentages $p$, used to reduce computational complexity.}
  \label{fig:permutation-experiment}
\end{figure}

Based on this observation, we vary the proportion $p$ to find an optimal trade-off between performance and efficiency. The full list of permutation used for each $p$ is detailed in Table~\ref{tab:permutation-coverage} in Appendix~\ref{sec:mad-table}. The impact of varying this percentage threshold on performance is illustrated in Figure~\ref{fig:permutation-experiment}.

An interesting finding is that using $50\%$ or $100\%$ of the permutations yields no significant difference in performance. The {\fone} remains relatively stable across this range. This empirically supports the idea that using only a subset of permutations can still yield high performance, as some permutations may produce similar log-probabilities. To balance computational cost and detection quality, we adopt $p = 50\%$ as the default threshold for the {\permutationr} method in the subsequent comparison.

\begin{figure}[htbp]
  \centering
  \includegraphics[width=1\linewidth]{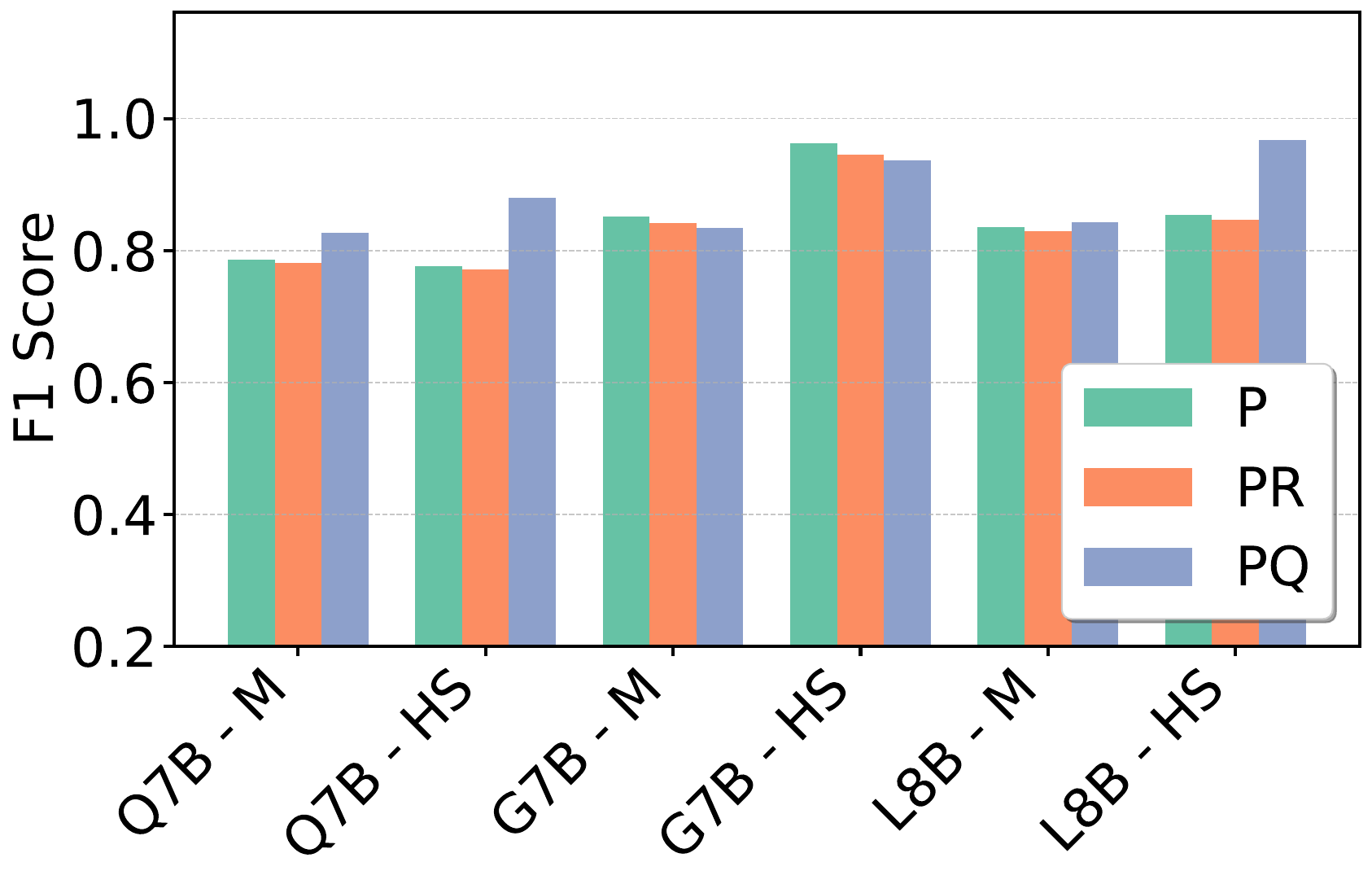}
  \caption{Performance of {\permutation}-based methods: the original (P), reduced variant {\permutationr} (PR), and quick variant {\permutationq} (PQ), evaluated on MMLU (M) and HellaSwag (HS). Model abbreviations: Q = Qwen, L = Llama, G = Gemma.\label{fig:permutation-Q-experiment}}
\end{figure}

\mypar{\texttt{Permutation-Q} Experiment}
We experiment {\permutationq} in six different model and dataset settings to observe its performance. We compare the result with {\permutationr} and the original {\permutation}. The {\fone} comparison is presented in Figure \ref{fig:permutation-Q-experiment}.

By using only two options per log-probability computation, {\permutationq} achieves competitive {\fone} scores and, in some settings, even outperforms both the original {\permutation} and {\permutationr} methods. These results highlight its ability to reduce complexity while potentially improving performance.

\subsection{Main Results}

\begin{table*}[t]
\centering
\resizebox{0.8\linewidth}{!}{%
\begin{tabular}{llcccccccc}
\toprule
\textbf{Benchmark} & \textbf{Model} & \textbf{S} & \textbf{P} & \textbf{PR} & \textbf{PQ} & \textbf{N} & \textbf{S+PQ} & \textbf{S+N} & \textbf{PQ+N} \\
\midrule
\multirow{4}{*}{MMLU} 
& Qwen-0.5B & $\DD55.68$ & $\DD82.78$ & $\DD82.12$ & $\DD86.63$ & \DD\underline{$88.23$} & $\DD79.16$ & $\DD79.84$ & $\DD85.47$ \\
& Qwen-7B   & $\DD56.88$ & $\DD78.64$ & $\DD78.12$ & \DD\underline{$82.68$} & $\DD81.89$ & $\DD78.79$ & $\DD78.37$ & $\DD80.59$ \\
& Gemma-7B  & $\DD68.59$ & \DD\underline{$85.10$} & $\DD84.14$ & $\DD83.45$ & $\DD82.87$ & $\DD77.12$ & $\DD75.95$ & $\DD80.32$ \\
& LLaMA-8B  & $\DD50.51$ & $\DD83.59$ & $\DD82.98$ & \DD\underline{$84.27$} & $\DD84.11$ & $\DD80.00$ & $\DD79.84$ & $\DD81.63$ \\
\midrule
\multirow{4}{*}{HellaSwag}
& Qwen-0.5B & $\DD60.86$ & $\DD81.13$ & $\DD80.73$ & $\DD94.94$ & \DD\underline{$99.83$} & $\DD80.61$ & $\DD82.99$ & $\DD96.46$ \\
& Qwen-7B   & $\DD67.92$ & $\DD77.67$ & $\DD77.10$ & $\DD87.93$ & \DD\underline{$96.01$} & $\DD73.48$ & $\DD75.79$ & $\DD95.67$ \\
& Gemma-7B  & $\DD71.04$ & $\DD96.20$ & $\DD94.50$ & $\DD93.69$ & \underline{$100.00$} & $\DD74.91$ & $\DD75.76$ & $\DD94.19$ \\
& LLaMA-8B  & $\DD68.71$ & $\DD85.40$ & $\DD84.70$ & $\DD96.77$ & \underline{$100.00$} & $\DD75.09$ & $\DD75.66$ & $\DD96.77$ \\
\bottomrule
\end{tabular}%
}
\caption{Detection performance ({\fone}) for various methods across different models and benchmarks. The methods are coded as follows: S = \texttt{Semi-half}, P = \texttt{Permutation}, PR = \texttt{Permutation-R}, PQ = \texttt{Permutation-Q}, and N = \texttt{N-Gram}. For combined methods (denoted by `+'), an instance is classified as Leakage if at least one of the methods detects it. \underline{Underlined} scores represent the best method among the model \& benchmark combinations.\label{tab:main-result}}
\end{table*}

\mypar{Detection Performance Across Methods with Tuned Thresholds}
Using the selected thresholds and configurations for the {\ngram} and {\permutationr} methods, along with other original approaches, we compare detection performance across eight evaluation settings. The results are presented in Table~\ref{tab:main-result}. Since each experiment uses a different subset of data, depending on the base model’s initial ability to answer the questions, the metrics should only be compared across detection methods within the same experiment, not across different models or benchmarks.

Across experiments, the {\ngram} method consistently achieves over $81\%$ {\fone} and outperforms other methods in Qwen-0.5B (MMLU) and all HellaSwag settings. \texttt{Permutation-Q} shows strong performance as well, outperforming other methods in Qwen-7B and LLaMA-8B, while the original {\permutation} achieves the best {\fone} in Gemma-7B. Overall, {\permutationq} performs competitively and often matches or exceeds the performance of {\ngram}.

Interestingly, combining multiple methods tends to increase {\rec} as more instances are flagged as `Leakage', but this often leads to a decrease in {\fone} (see Table~\ref{tab:detection-results} in Appendix~\ref{sec:full-comparison}), likely due to a rise in false positives and lower {\preci}.
A closer look at the HellaSwag results reveals that {\ngram} almost perfectly detects `Leakage' across all settings. This may be attributed to the nature of the HellaSwag task, which involves predicting the most coherent continuation of a given context. This objective closely aligned with how {\ngram} generates options based on prefix patterns. It is also possible that {\ngram} benefits from the continual pretraining objective, which focuses on next-token prediction. Regardless of the cause, {\ngram} remains highly effective and competitive. Furthermore, since it does not require access to model logits, it can be applied to closed-weight models. For these reasons, we adopt {\ngram} as the primary detection method for both MMLU and HellaSwag. 

\begin{figure}[t]
  \centering
  \includegraphics[width=1\linewidth]{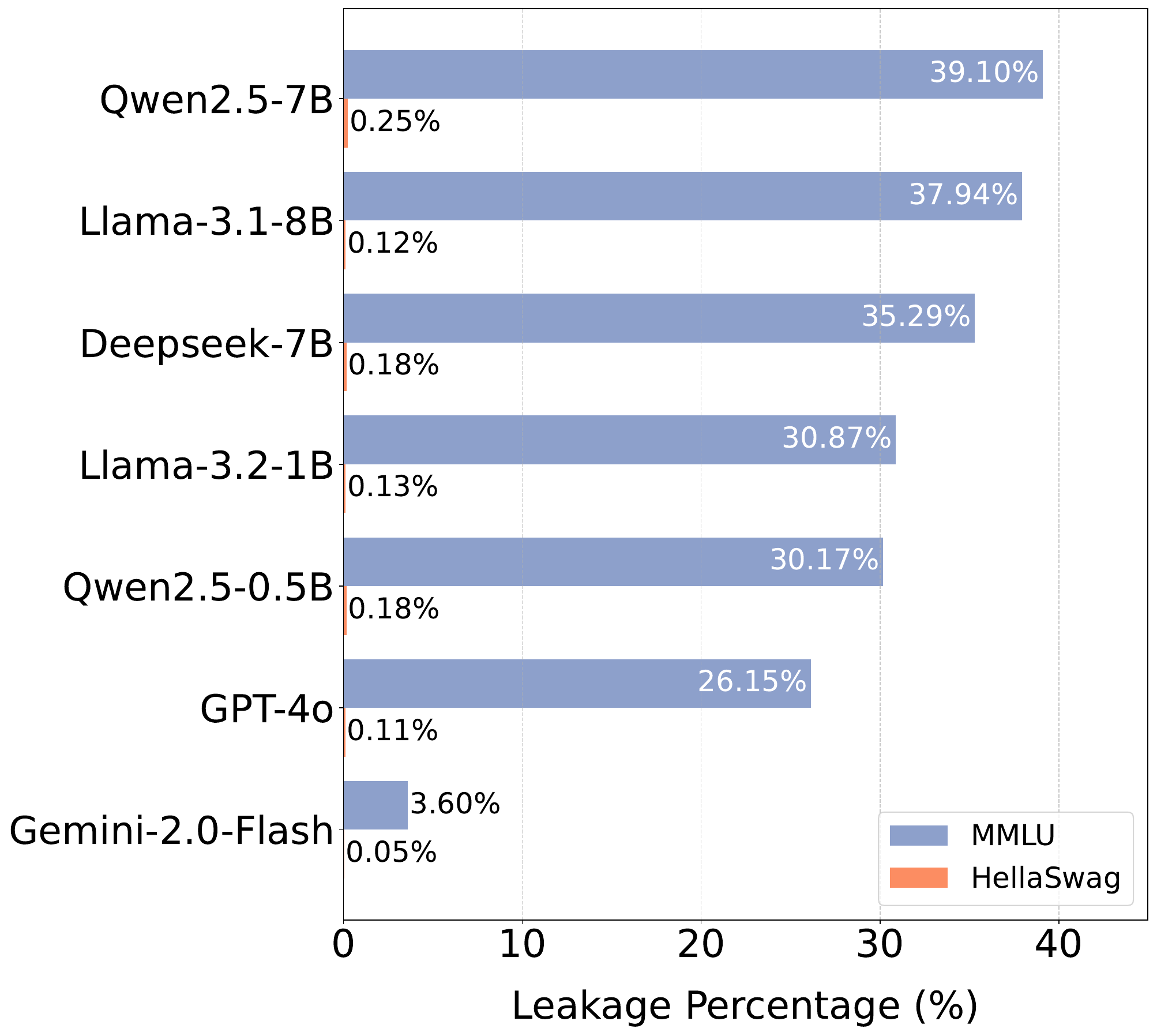}
  \caption{Data leakage rates for each model on MMLU and HellaSwag using the {\ngram} method.}
  \label{fig:leakage-percentages}
\end{figure}

\mypar{Leakage Detection Results on Full Benchmarks}
After applying the {\ngram} method to the full MMLU and HellaSwag datasets, we identified several instances flagged as `Leakage'. Figure~\ref{fig:leakage-percentages} illustrates the proportion of detected leakage across different models. Qwen2.5-7B shows a relatively higher tendency for potential leakage in both benchmarks, followed by LLaMA-3.1-8B on MMLU and Qwen2.5-0.5B on HellaSwag. These observations align with the findings of \citet{ni2024training}, who also highlighted potential risks of leakage in the Qwen model family, despite using a different methodology. We additionally tested DeepSeek \cite{liu2024deepseek} and Gemini \cite{team2023gemini} models: DeepSeek ranks third with 35\% on MMLU and 0.17\% on HellaSwag, while Gemini-2.0-Flash exhibits minimal indications of leakage across both datasets.

We observe that MMLU shows a higher potential for leakage across models compared to HellaSwag, likely due to its widespread use in NLP research, making its content more likely to appear in training data. Additionally, the {\ngram} method is sensitive to the length of the option text, as it generates tokens sequentially to match the reference, resulting in slower detection for longer options. In contrast, {\semihalf} and {\permutation}-based methods require only a single inference step per instance, making them more efficient.

\mypar{Leakage $\neq$ Model Understanding}
We observe that models do not always correctly answer leaked instances, both in our leakage simulation and full-benchmark evaluations. As illustrated in Table~\ref{tab:leaked-score}, models frequently fail to provide correct responses to flagged examples, indicating that memorizing input sequences does not equate to genuine understanding or reasoning. Nonetheless, instances encountered during pretraining are generally more likely to be answered correctly. This discrepancy may arise from a misalignment between training and evaluation objectives: while continual pretraining does not aim to identify the most likely answer among multiple choices, evaluation typically depends on comparing the log-likelihoods of each option.

\begin{table}[t]
\centering
\resizebox{0.8\linewidth}{!}{%
    \begin{tabular}{lc}
    \toprule
    \textbf{Model} & \textbf{Accuracy on Leaked Set} \\
    \midrule
    Deepseek-7B & $39.06\%$ \\
    Gemini-2.0-Flash & $95.65\%$ \\
    GPT-4o & $87.09\%$ \\
    LLaMA-3.1-8B & $51.19\%$ \\
    Qwen2.5-7B & $64.12\%$ \\
    \bottomrule
    \end{tabular}
}
\caption{Accuracy scores of models on MMLU instances detected as leakage by {\ngram}.\label{tab:leaked-score}}
\end{table}

Despite this disconnect, computing the log-likelihood of each option (A–D) remains the standard for evaluating multiple-choice questions in LLM. However, since we lack visibility into how each model was exposed to these benchmark—such as whether answer keys were included during pretraining—we propose two complementary definitions of leakage:
(1) Strong leakage: the instance is detected as `Leakage' and the model answers it correctly.
(2) Weak leakage: the instance is detected as `Leakage', regardless of the model’s answer.

\mypar{Benchmark Reduction Under Strong Leakage Definition} By the strong definition of leakage, we remove any instance flagged as `Leakage' by the {\ngram} method in at least one LLM. This results in the removal of $6{,}547$ out of $14{,}042$ instances ($46.6\%$) from MMLU, and $38$ out of $10{,}042$ instances ($0.38\%$) from HellaSwag. 

\begin{figure}[htbp]
  \centering
  \includegraphics[width=1\linewidth]{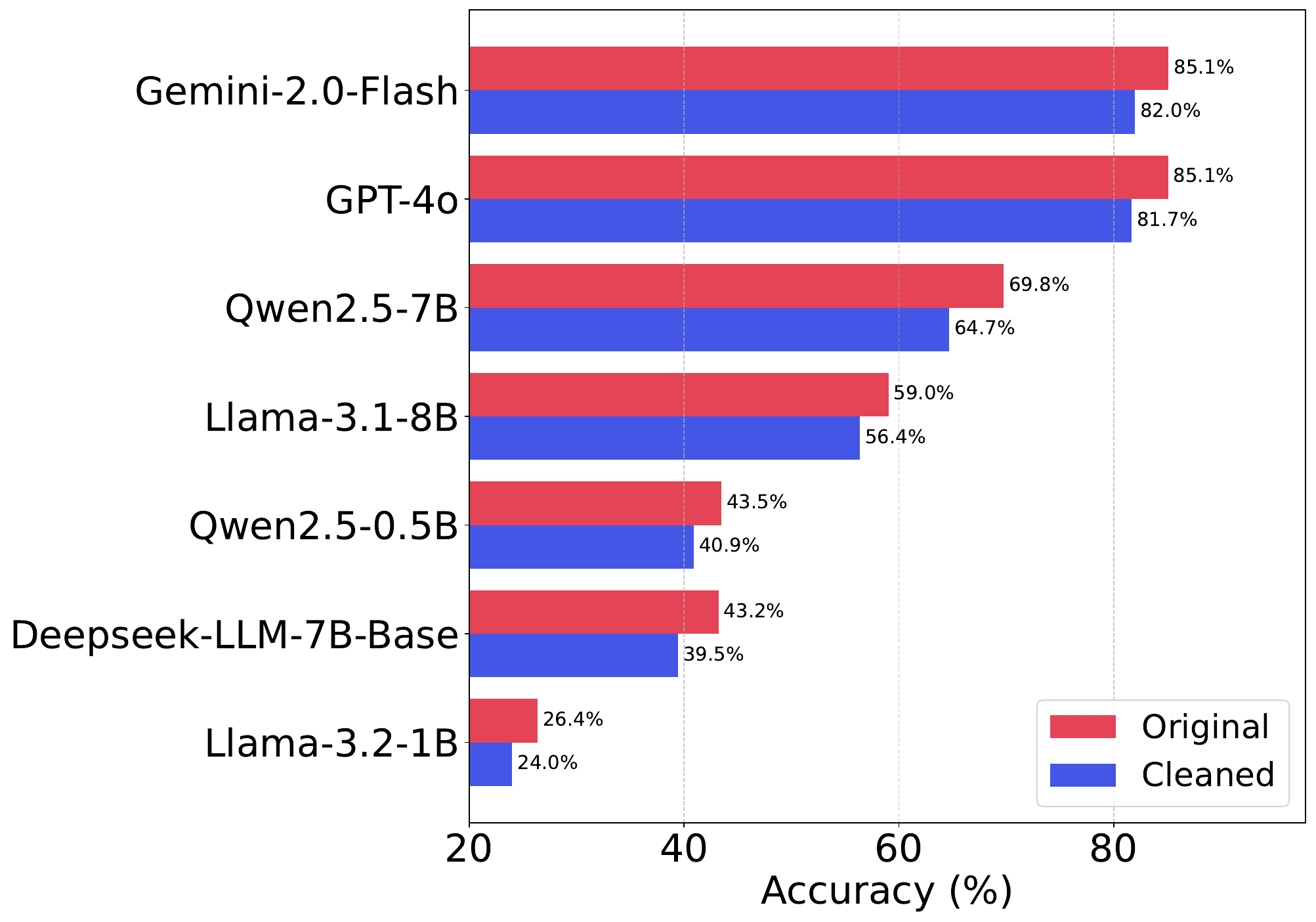}
  \caption{Comparison of model performance on original vs. cleaned MMLU benchmark based on strong definition of leakage.}
  \label{fig:benchmark_comparison}
\end{figure}

Figure~\ref{fig:benchmark_comparison} compares model performance on the original and cleaned versions of MMLU.\footnote{HellaSwag is excluded due to the minimal number of leaked instances (0.38\%) and the absence of notable performance differences. GPT-4o remains the top-performing model on both versions.} On the cleaned MMLU benchmark, Gemini-2.0-Flash achieves the highest performance among all evaluated models, followed closely by GPT-4o. While the relative ranking across the evaluated models remains largely consistent even after removing 46\% of potentially leaked instances, we note that the models differ in size. Therefore, performance shifts could be more pronounced when comparing models within the same parameter scale. To explore this further, we analyze accuracy drops by subject and subject category within the same model in Section~\ref{sec:analysis}.

We also tested the LLMs' performance with the weak leakage definition. Since model accuracy on leaked instances is not $100\%$, removing these instances reduces the number of incorrectly answered examples, resulting in higher accuracy for some models. However, this only affects the performance ranking on the MMLU dataset, where GPT-4o slightly surpasses Gemini 2.0 Flash to become the top-performing model. The ranking positions for the other models remain unchanged.
\section{Analysis}
\label{sec:analysis}
\subsection{Performance Varies in Specific Subjects}

\begin{figure}[htbp]
\centering
\includegraphics[width=1\linewidth]{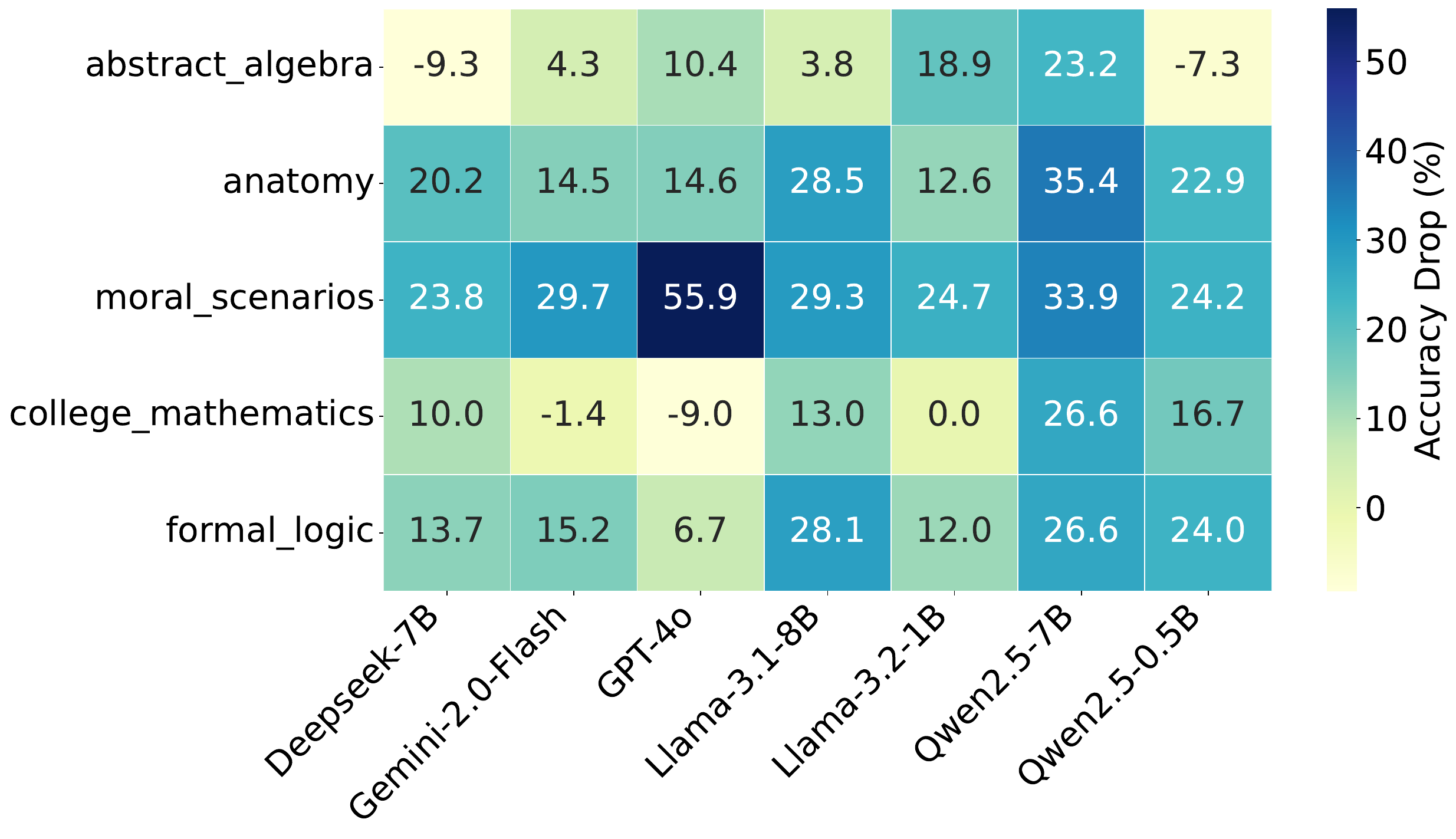}
\caption{Performance drops in selected MMLU subjects for each model. Qwen-7B shows the largest accuracy drop in these subjects.}
\label{fig:diff_subject_selected}
\end{figure}

Referring to the strong version of leakage definition, we analyze performance changes across specific MMLU subjects. Figure~\ref{fig:diff_subject_selected} presents the percentage drop in accuracy after removing potentially leaked instances, which highlights subjects with relatively large performance declines. In particular, model performance on the \textit{Anatomy} subject drops substantially, with Qwen-7B showing the largest decrease ($35.4\%$).

Meanwhile, \textit{Moral Scenarios} contains the highest number of detected leaked instances. However, as observed in previous experiments, this subject yields a low {\fone}. Upon further inspection, we find this is likely due to repetitive option templates used across all questions, such as ``\emph{Not Wrong, Not Wrong; Not Wrong, Wrong; Wrong, Not Wrong; Wrong, Wrong}'' or ``\emph{True, False}.'' These patterns may increase the chance that the model generates the correct option based solely on surface similarity, leading to false positives under the {\ngram} detection method.

We also observe a noticeable accuracy decline in \textit{Formal Logic} after data cleaning. This suggests that part of the model’s original strong performance in this subject could be attributed to memorization rather than genuine reasoning ability. The cleaned results provide a more realistic reflection of the models' logical reasoning skills.

\begin{figure}[htbp]
\centering
\includegraphics[width=1\linewidth]{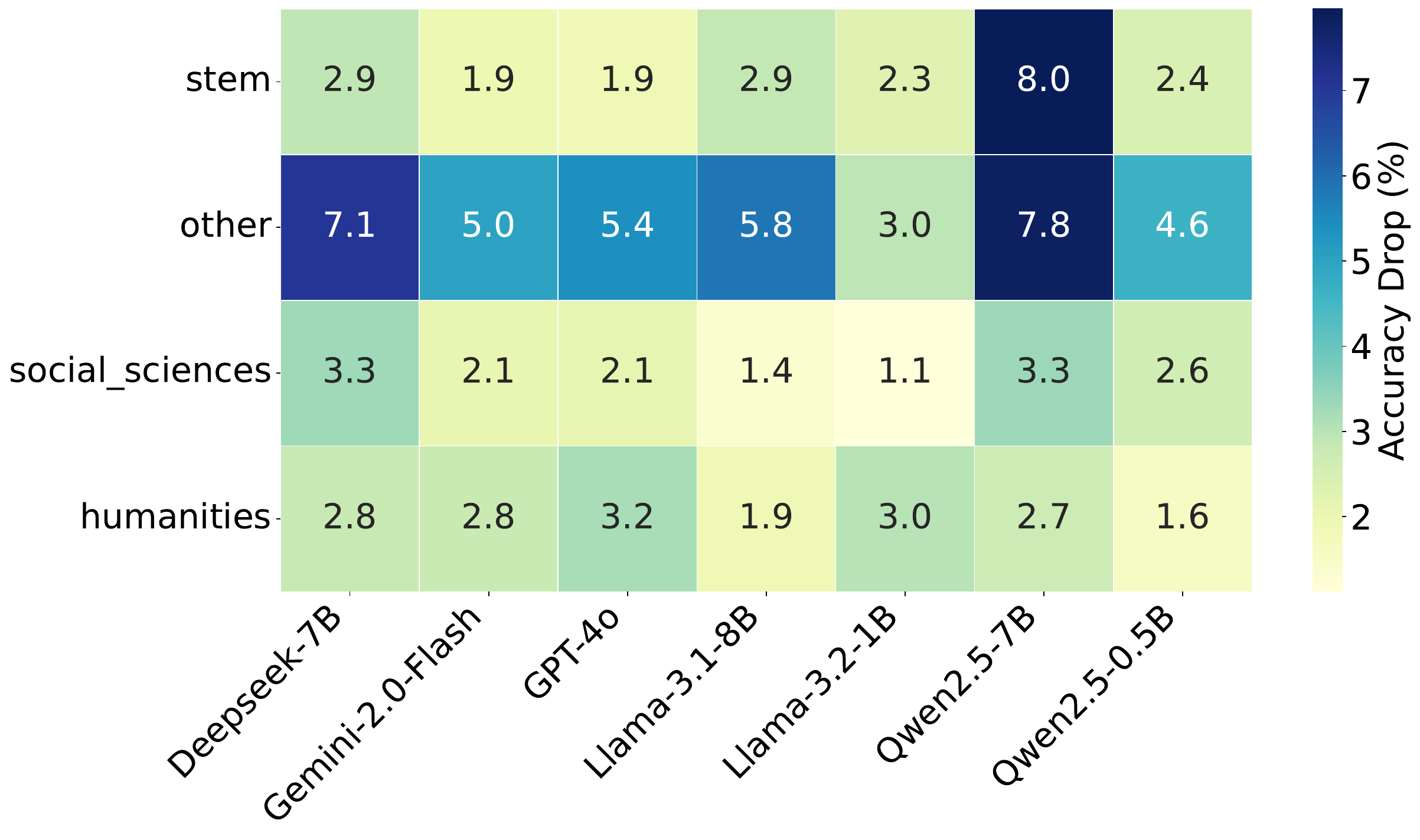}
\caption{Performance drops by subject group for each model. Qwen-7B shows a marked drop in the STEM group. The 'Other' category exhibits the most performance decline overall.}
\label{fig:diff_group}
\end{figure}

Across broader subject groups (Figure~\ref{fig:diff_group}), Qwen-7B’s performance in the \textit{STEM} group appears more affected, with an observed accuracy drop of up to $8\%$. All models also experience a decline in the \textit{Other} category, with Qwen-7B again showing the most pronounced decrease.

\subsection{Detection Methods: Pros \& Cons}
As shown in Table~\ref{tab:main-result}, the {\ngram} method consistently achieves the highest {\fone}, followed closely by \texttt{Permutation-Q}. However, each method has its own strengths and weaknesses. Table~\ref{tab:pros-cons-restructured} in Appendix~\ref{sec:pros-cons-comparison} summarizes the strengths and limitations of each method that focusing on computational cost and leakage detection effectiveness, to inform their use in different scenarios.
\section{Conclusion}

In this work, we simulate leakage using three methods—\semihalf, \permutation, and \ngram—and introduce a simplified variant, \permutationq, which uses only two options and achieves strong performance across several settings. Our results identify {\permutationq} and {\ngram} as the most effective detection methods, with Qwen-7B showing a high risk of leakage, especially in MMLU STEM subjects, where accuracy drops by up to $8\%$ after filtering. We also observe consistent accuracy reductions across all models once leaked instances are removed. These findings highlight the distinction between memorization and true understanding, reinforcing the need to apply leakage detection before evaluation to ensure that test data remains clean and that the reported results reflect genuine generalization.

\section*{Limitations}
Detection methods tend to yield a high false positive rate on moral scenario subjects. This is likely due to the repetitive structure of moral scenario questions in MMLU, which remains consistent across instances. In general, this issue arises in questions that use common option formats without referencing a specific domain topic, such as ``\emph{True; False}'' or ``\emph{First; Second; Third; Fourth}''. As a result, when the model encounters a new question with a similar option structure, it may mistakenly flag it as a leakage instance. We consider this a limitation of our approach and encourage future work to explore more robust strategies for detecting leakage in cases involving repeated option sentences or generic question formats.

While simulating continual pre-training (CPT), we recognize that using a single benchmark as the sole training corpus does not fully capture real-world LLM training. However, to better reflect practical scenarios, we also performed continual pre-training with the benchmark mixed with additional random data. Detection performance stayed consistent in both setups, and we observed no significant drop in the performance.

\section*{Acknowledgements}
This research was supported by funding and resources from the Mohammed bin Zayed University of Artificial Intelligence (MBZUAI), whose contribution is gratefully acknowledged. The authors also extend their appreciation to the Tokopedia-UI AI Center of Excellence for providing GPU compute resources, and to the Faculty of Computer Science, Universitas Indonesia, for its comprehensive institutional support and facilitation throughout the research process.

\section*{Ethics Statements}
This work investigates potential data contamination in benchmark evaluations of large language models. While our methods aim to surface instances likely seen during training, we make no claims about the intentional inclusion of benchmark data or misconduct by model developers. Our findings are intended to support more reliable and transparent evaluation practices. We discourage the use of these results to make unfounded accusations against any specific model or organization.

\bibliography{custom}
\bibliographystyle{acl_natbib}

\appendix
\clearpage
\section{Detailed Permutation Method}
\label{sec:detailed-permutation}
For each instance, we compute the log-probability score for all possible permutations of the answer options. Let $n$ be the number of answer options, resulting in $n!$ possible permutations. For each permutation, denoted as $\pi = (o_{\pi(1)}, o_{\pi(2)}, \ldots, o_{\pi(n)})$, we construct the sequence $[q, o_{\pi(1)}, o_{\pi(2)}, \ldots, o_{\pi(n)}]$, where $q$ is the question and $o_{\pi(i)}$ are the permuted answer options. We then tokenize this sequence into $x = (x_1, x_2, \ldots, x_T)$, where $T$ is the total number of tokens.

Given a language model $\mathcal{M}$, we compute the log-probability score starting from the first token of the first answer option. The score is calculated as:
\[\text{Score}(x)=\sum_{i=i^*}^{T} \log P(x_i \mid x_{<i}; \mathcal{M}) \, ,
\]
\noindent
where $P(x_i \mid x_{<i}; \mathcal{M})$ is the conditional probability assigned to token $x_i$ given its preceding context; and $i^*$ marks the token index where the first option sentence begins. This scoring function captures how likely the model is to continue generating a specific option sequence, conditioned on the prompt. If the original order (A-B-C-D) has the maximum log-probability across other orders, we consider the model to memorize that version more and the instance as `Leakage', otherwise not.

\section{Permutation-Q Algorithm}
\label{sec:permutation-q-algorithm}
Algorithm \ref{alg:permutation-Q} details our refined {\permutationq} procedure. Its time complexity is lower than that of the original {\permutation} method because the factorial term ($O(n!)$) is replaced by a quadratic factor ($O(n^{2})$).

\begin{algorithm}[htbp]
\caption{\texttt{Permutation-Q} Detection Method\label{alg:permutation-Q}}
\textbf{Input:} Data $x = [q, o_1, o_2, ..., o_n]$; Model $\mathcal{M}$ \\
\textbf{Output:} ``L'' (Leaked) or ``NL'' (Not Leaked)
\begin{algorithmic}[1]
\State \textit{\color{blue}\# Generate all pairs of options}
\State $P \leftarrow \texttt{Permute}(\{o_1, o_2, ..., o_n\}, 2)$
\State \textit{\color{blue}\# Initialize empty score list}
\State $\texttt{scores} \leftarrow [\DD]$
\For{\textnormal{each pair $(o_i, o_j) \in P$}}
    \State \textit{\color{blue}\# Construct prompt sequence}
    \State $\texttt{seq} \leftarrow [q, o_i, o_j]$
    \State \textit{\color{blue}\# Compute log-probability}
    \State $\texttt{scores} \textrm{ += } [log(P(\texttt{seq}; \mathcal{M}))]$
\EndFor
\State \textit{\color{blue}\# Construct original correct sequence}
\State $\texttt{seq}' \leftarrow [q, o_1, o_2]$ 
\State \textit{\color{blue}\# Has the highest log-probability?}
\If{$log (P(\texttt{seq}'; \mathcal{M})) = \textrm{max}(\texttt{scores})$}
    \State \Return ``L'' \Comment{{\color{blue}Leaked}}
\Else
    \State \Return ``NL'' \Comment{{\color{blue}Not Leaked}}
\EndIf
\end{algorithmic}
\end{algorithm}

\section{N-gram Algorithm}
\label{sec:ngram-algorithm}

For each input instance $x = [q, o_1, o_2, \ldots, o_n]$, the model $\mathcal{M}$ is asked to generate each option $o_i$ ($i \leq n$), using the question $q$ and the previous options $o_1$ to $o_{i-1}$ as context. The generated output $\hat{o}_i$ is then compared to the original $o_i$ using a \texttt{ROUGE-L} score \cite{lin-2004-rouge}. If the similarity score is above a threshold $t=0.75$ (based on \citet{xu2024benchmarking}), we consider the option as replicated. We count how many options are replicated in this way and calculate the ratio over the total number of options. If this ratio is higher than a threshold $T$, we mark the instance $x$ as contaminated for model $\mathcal{M}$. A full, detailed algorithm is presented in Algorithm \ref{alg:ngram-detection}.

The threshold used for this detection method decides the sensitivity level. If we use $T=0.25$, meaning an instance is labeled as `Leakage' if at least one of its options is generated with high similarity to the ground truth, this is intended to capture as many suspicious instances as possible. The smaller the threshold $T$ is, the more sensitive it gets to detect contamination.
In this study, we further explore the effect of varying the threshold $T$. Since both MMLU and HellaSwag benchmarks contain four options per question, we experiment with $T=\{0.00, 0.25, 0.5, 0.75, 1.00\}$, where each value represents the minimum proportion of similar options required to consider a question contaminated. 

\begin{algorithm}[htbp]
\caption{\texttt{N-Gram} Detection Method\label{alg:ngram-detection}}
\textbf{Input:} Data $x = [q, o_1, o_2, ..., o_n]$; Model $\mathcal{M}$; Similarity threshold $t$; Leakage threshold $T$\\
\textbf{Output:} ``L'' (Leaked) or ``NL'' (Not Leaked)
\begin{algorithmic}[1]
\State \textit{\color{blue}\# Initialize count}
\State \texttt{count} $\leftarrow$ $0$
\For{ $i$ $=$ $1$ to $n$ }
    \State \textit{\color{blue}\# Construct prompt}
    \State \texttt{prompt} $\leftarrow$ $[q, o_1, o_2, ..., o_{i-1}]$
    
    \State \textit{\color{blue}\# Generate prediction}
    \State $\hat{o_i} \leftarrow \mathcal{M}(\texttt{prompt})$
    
    \State \textit{\color{blue}\# Compute similarity score}
    \State \texttt{score} $\leftarrow$ \texttt{ROUGE-L}($\hat{o_i}$, $o_i$)
    
    \If{ \texttt{score} $\geq$ t }
        \State \texttt{count} $\leftarrow$ \texttt{count} + 1
    \EndIf
\EndFor
\State \texttt{ratio} $\leftarrow$ \texttt{count} $/$ $n$
\If{ \texttt{ratio} $\geq$ $T$ }
    \State \Return ``L'' \Comment{{\color{blue}Leaked}}
\Else
    \State \Return ``NL'' \Comment{{\color{blue}Not Leaked}}
\EndIf
\end{algorithmic}
\end{algorithm}

\section{Average Ranking Ordering Across Permutations}
\label{sec:mad-table}

After we compute Mean Absolute Difference (MAD) for each models, we average the ranking to get the order of most similar permutations, shown in Table~\ref{tab:rank-diff}. These order reflect how similarly a model responds to different answer orderings.

To decide which permutation used for a certain percentage $p$, we iteratively eliminate one permutation from each highly similar pair. Specifically, we remove the second permutation in the pair, assuming its behavior is already well-represented by the first. This process continues until the desired number of permutations, determined by a percentage threshold $p$, remains.

The final set of retained permutations for each threshold level $p$ used in the {\permutationr} experiment is detailed in Table~\ref{tab:permutation-coverage}. This approach ensures that we retain a diverse set of permutations while minimizing redundant evaluation.

\begin{table}[t]
\centering
\begin{tabular}{lc}
\toprule
\textbf{Permutation Pair} & \textbf{Average Rank} \\
\midrule
ACBD - ACDB & \DD$2.67$ \\
CDAB - CDBA & \DD$3.67$ \\
BACD - BCAD & \DD$4.67$ \\
CADB - CDAB & \DD$7.33$ \\
ACDB - ADBC & $10.33$ \\
DBAC - DBCA & $10.67$ \\
BACD - BADC & $15.00$ \\
DCAB - DCBA & $17.67$ \\
ACBD - ADBC & $17.67$ \\
BDAC - BDCA & $18.00$ \\
CBDA - CDBA & $19.00$ \\
ADBC - ADCB & $19.33$ \\
DBCA - DCBA & $20.67$ \\
CBAD - CBDA & $21.67$ \\
CABD - CBAD & $24.33$ \\
CADB - CDBA & $27.00$ \\
ACDB - ADCB & $27.67$ \\
BADC - BCAD & $28.33$ \\
DBCA - DCAB & $33.00$ \\
DBAC - DCAB & $34.00$ \\
ACBD - BACD & $35.00$ \\
ACDB - BACD & $36.67$ \\
CADB - CBDA & $37.00$ \\
ADCB - DABC & $41.00$ \\
\bottomrule
\end{tabular}
\caption{Top-24 average rank between permutation pairs in Qwen-7B, LLaMA-8B, and Gemma-7B, sorted in increasing order. Lower average rank indicate higher similarity.\label{tab:rank-diff}}
\end{table}

\begin{table*}[htbp]
\centering
\begin{tabular}{c p{12cm}} 
\toprule
\textbf{Percentage ($p$)} & \textbf{Permutations Used} \\
\midrule
\DD$0$ & ABCD \\
\midrule
\DD$10$ & ABCD, ABDC \\
\midrule
\DD$20$ & ABCD, ABDC, ACBD, CABD \\
\midrule
\DD$30$ & ABCD, ABDC, ACBD, BCDA, CABD, CADB, DBAC \\
\midrule
\DD$40$ & ABCD, ABDC, ACBD, BCDA, BDAC, CABD, CADB, DACB, DBAC \\
\midrule
\DD$50$ & ABCD, ABDC, ACBD, BACD, BCDA, BDAC, CABD, CADB, DABC, DACB, DBAC, DCAB \\
\midrule
\DD$60$ & ABCD, ABDC, ACBD, BACD, BCDA, BDAC, CABD, CADB, CBAD, CBDA, DABC, DACB, DBAC, DCAB \\
\midrule
\DD$70$ & ABCD, ABDC, ACBD, ADCB, BACD, BCDA, BDAC, BDCA, CABD, CADB, CBAD, CBDA, DABC, DACB, DBAC, DCAB \\
\midrule
\DD$80$ & ABCD, ABDC, ACBD, ADCB, BACD, BADC, BCDA, BDAC, BDCA, CABD, CADB, CBAD, CBDA, DABC, DACB, DBAC, DBCA, DCAB, DCBA \\
\midrule
\DD$90$ & ABCD, ABDC, ACBD, ADBC, ADCB, BACD, BADC, BCDA, BDAC, BDCA, CABD, CADB, CBAD, CBDA, CDAB, DABC, DACB, DBAC, DBCA, DCAB, DCBA \\
\midrule
$100$ & ABCD, ABDC, ACBD, ACDB, ADBC, ADCB, BACD, BADC, BCAD, BCDA, BDAC, BDCA, CABD, CADB, CBAD, CBDA, CDAB, CDBA, DABC, DACB, DBAC, DBCA, DCAB, DCBA \\
\bottomrule
\end{tabular}
\caption{Permutations used at each $p$ percentage level for {\permutationr}.\label{tab:permutation-coverage}}
\end{table*}

\section{Detailed Model Used in Experiment}
\label{sec:model-setting}

Table~\ref{tab:model-sources} lists the LLMs used in our experiments. Each model is evaluated on both the MMLU and HellaSwag benchmarks. The models include Qwen \citep{qwen2025qwen25technicalreport}, Gemma \citep{gemma2}, and LLaMA \citep{touvron2023llamaopenefficientfoundation}. All models are accessed via Hugging Face \citep{wolf2020huggingfacestransformersstateoftheartnatural}.

\begin{table*}[htbp]
\centering
\begin{tabular}{@{}l >{\raggedleft\arraybackslash}p{0.5\textwidth}@{}}
\toprule
\textbf{Model (\#Parameter)} & \textbf{Source} \\
\midrule
Qwen (0.5B) & \texttt{Qwen/Qwen2.5-0.5B} \\
Qwen (7B)   & \texttt{Qwen/Qwen2.5-7B} \\
Gemma (7B)  & \texttt{google/gemma-7b} \\
LLaMA (8B)  & \texttt{meta-llama/Llama-3.1-8B} \\
\bottomrule
\end{tabular}
\caption{Model sources used in the experiments. All models are accessed via Hugging Face \citep{wolf2020huggingfacestransformersstateoftheartnatural}.}
\label{tab:model-sources}
\end{table*}

\section{Complete Detection Methods Performance Comparison}
\label{sec:full-comparison}
Table~\ref{tab:detection-results} presents the full comparison of detection method performance in different model and evaluation settings. This table provides a more detailed overview of each method's sensitivity in detecting leakage ({\rec}), as well as its effectiveness in identifying true leakage while minimizing false positives ({\preci}).
\begin{table*}[htbp]
\centering
\resizebox{\textwidth}{!}{%
\begin{tabular}{llcccccccc}
\toprule
\multirow{2}{*}{\textbf{Method}} & \multirow{2}{*}{\textbf{Metric}} & \multicolumn{4}{c}{\textbf{MMLU}} & \multicolumn{4}{c}{\textbf{HellaSwag}} \\
\cmidrule(lr){3-6} \cmidrule(lr){7-10}
& & \textbf{Qwen-0.5B} & \textbf{Qwen-7B} & \textbf{Gemma-7B} & \textbf{LLaMA-8B} & \textbf{Qwen-0.5B} & \textbf{Qwen-7B} & \textbf{Gemma-7B} & \textbf{LLaMA-8B} \\
\midrule
\multirow{3}{*}{S}
& Recall    & \DD$50.67$ & \DD$51.00$ & \DD$71.33$ & \DD$41.00$ & \DD$61.67$ & \DD$84.00$ & \DD$90.33$ & \DD$86.00$ \\
& Precision & \DD$61.79$ & \DD$64.29$ & \DD$66.05$ & \DD$65.78$ & \DD$60.06$ & \DD$57.01$ & \DD$58.53$ & \DD$57.21$ \\
& F1-Score  & \DD$55.68$ & \DD$56.88$ & \DD$68.59$ & \DD$50.51$ & \DD$60.86$ & \DD$67.92$ & \DD$71.04$ & \DD$68.71$ \\
\midrule
\multirow{3}{*}{P}
& Recall    & \DD$87.33$ & \DD$88.33$ & \DD$99.00$ & \DD$97.67$ & \DD$71.67$ & \DD$66.67$ & \DD$97.00$ & \DD$78.00$ \\
& Precision & \DD$78.68$ & \DD$70.86$ & \DD$74.62$ & \DD$73.07$ & \DD$93.48$ & \DD$93.02$ & \DD$95.41$ & \DD$94.35$ \\
& F1-Score  & \DD$82.78$ & \DD$78.64$ & \DD$\textbf{85.10}$ & \DD$83.59$ & \DD$81.13$ & \DD$77.67$ & \DD$\textbf{96.20}$ & \DD$85.40$ \\
\midrule
\multirow{3}{*}{PR}
& Recall    & \DD$88.00$ & \DD$88.67$ & \DD$99.00$ & \DD$98.33$ & \DD$74.00$ & \DD$67.33$ & \DD$97.33$ & \DD$79.33$ \\
& Precision & \DD$76.97$ & \DD$69.82$ & \DD$73.15$ & \DD$71.78$ & \DD$88.80$ & \DD$90.18$ & \DD$91.82$ & \DD$90.84$ \\
& F1-Score  & \DD$82.12$ & \DD$78.12$ & \DD$84.14$ & \DD$82.98$ & \DD$80.73$ & \DD$77.10$ & \DD$94.50$ & \DD$84.70$ \\
\midrule
\multirow{3}{*}{PQ}
& Recall    & \DD$99.33$ & \DD$98.67$ & $100.00$ & $100.00$ & \DD$97.00$ & \DD$85.00$ & \DD$99.00$ & $100.00$ \\
& Precision & \DD$76.80$ & \DD$71.15$ & \DD$71.60$ & \DD$72.82$ & \DD$92.97$ & \DD$91.07$ & \DD$88.92$ & \DD$93.75$ \\
& F1-Score  & \DD$86.63$ & \DD$\textbf{82.68}$ & \DD$83.45$ & \DD$\textbf{84.27}$ & \DD$94.94$ & \DD$87.93$ & \DD$93.69$ & \DD$96.77$ \\
\midrule
\multirow{3}{*}{N}
& Recall    & \DD$98.67$ & \DD$98.00$ & $100.00$ & \DD$99.67$ & \DD$99.67$ & \DD$92.33$ & $100.00$ & $100.00$ \\
& Precision & \DD$79.78$ & \DD$70.33$ & \DD$70.75$ & \DD$72.75$ & $100.00$ & $100.00$ & $100.00$ & $100.00$ \\
& F1-Score  & \DD$\textbf{88.23}$ & \DD$81.89$ & \DD$82.87$ & \DD$84.11$ & \DD$\textbf{99.83}$ & \DD$\textbf{96.01}$ & $\textbf{100.00}$ & $\textbf{100.00}$ \\
\midrule
\multirow{3}{*}{S + PQ}
& Recall    & $100.00$ & \DD$99.67$ & $100.00$ & $100.00$ & \DD$97.67$ & \DD$97.00$ & $100.00$ & $100.00$ \\
& Precision & \DD$65.50$ & \DD$65.14$ & \DD$62.76$ & \DD$66.67$ & \DD$68.62$ & \DD$59.15$ & \DD$59.88$ & \DD$60.12$ \\
& F1-Score  & \DD$79.16$ & \DD$78.79$ & \DD$77.12$ & \DD$80.00$ & \DD$80.61$ & \DD$73.48$ & \DD$74.91$ & \DD$75.09$ \\
\midrule
\multirow{3}{*}{S + N}
& Recall    & \DD$99.00$ & \DD$99.67$ & $100.00$ & \DD$99.67$ & $100.00$ & \DD$99.67$ & $100.00$ & $100.00$ \\
& Precision & \DD$66.89$ & \DD$64.58$ & \DD$61.22$ & \DD$66.59$ & \DD$70.92$ & \DD$61.15$ & \DD$60.98$ & \DD$60.85$ \\
& F1-Score  & \DD$79.84$ & \DD$78.37$ & \DD$75.95$ & \DD$79.84$ & \DD$82.99$ & \DD$75.79$ & \DD$75.76$ & \DD$75.66$ \\
\midrule
\multirow{3}{*}{PQ + N}
& Recall    & $100.00$ & \DD$99.67$ & $100.00$ & $100.00$ & $100.00$ & \DD$99.33$ & $100.00$ & $100.00$ \\
& Precision & \DD$74.63$ & \DD$67.65$ & \DD$67.11$ & \DD$68.97$ & \DD$93.17$ & \DD$92.26$ & \DD$89.02$ & \DD$93.75$ \\
& F1-Score  & \DD$85.47$ & \DD$80.59$ & \DD$80.32$ & \DD$81.63$ & \DD$96.46$ & \DD$95.67$ & \DD$94.19$ & \DD$96.77$ \\
\bottomrule
\end{tabular}%
}
\caption{Detection performance (Recall, Precision, and F1-score) for various methods across different models and benchmarks. \textbf{Bold} scores represent the best F1-score among several leakage detection methods. The methods are coded as follows: S = Semi-half, P = Permutation, PR = Permutation-R, PQ = Permutation-Q, and N = N-Gram. \label{tab:detection-results}}
\end{table*}

\section{Experiment with Instruct Model}
Besides the experiment with only using base models, we also apply the same procedure to an instruct model. Table~\ref{tab:instruct-results} shows the performance comparison between base and instruct models. We can see that the {\fone} in the instruct model mostly produces a larger score than the base model.
\label{sec:instruct-comparison}
\begin{table*}[htbp]
\centering
\begin{tabular}{lcccc}
\toprule
\multirow{2}{*}{\textbf{Method}} & \multicolumn{2}{c}{\textbf{MMLU}} & \multicolumn{2}{c}{\textbf{HellaSwag}} \\
\cmidrule(lr){2-3} \cmidrule(lr){4-5}
& \textbf{Base} & \textbf{Instruct} & \textbf{Base} & \textbf{Instruct} \\
\midrule
Semi-half & \DD$55.68$ & \DD$76.67$ & \DD$60.86$ & \DD$69.92$ \\
Permutation & \DD$82.78$ & \DD$87.04$ & \DD$81.13$ & \DD$87.78$ \\
Permutation-R & \DD$82.12$ & \DD$85.84$ & \DD$80.73$ & \DD$86.74$ \\
Permutation-Q & \DD$86.63$ & \DD$87.55$ & \DD$94.94$ & \DD$95.01$ \\
N-Gram & \DD$88.23$ & \DD$88.79$ & \DD$99.83$ & $100.00$ \\
Semi-half + Permutation-Q & \DD$79.16$ & \DD$80.92$ & \DD$80.61$ & \DD$80.65$ \\
Semi-half + N-Gram & \DD$79.84$ & \DD$81.87$ & \DD$82.99$ & \DD$82.64$ \\
Permutation-Q + N-Gram & \DD$85.47$ & \DD$85.67$ & \DD$96.46$ & \DD$95.85$ \\
\bottomrule
\end{tabular}
\caption{{\fone} comparison across different detection methods between Qwen 0.5B base and instruct models on MMLU and HellaSwag datasets.\label{tab:instruct-results}}
\end{table*}

\section{Detailed Detection Methods' Pros \& Cons}
\label{sec:pros-cons-comparison}
Table \ref{tab:pros-cons-restructured} summarizes the advantages and limitations of the leakage-detection methods discussed in this paper across four criteria: computation time, detection effectiveness, risk of misclassification, and compatibility with closed-weight models. Compatibility with closed-weight models is crucial because many state-of-the-art LLMs do not release their weights, making certain detection methods unusable for their evaluation.

\begin{table*}[ht]
\centering
\small
\begin{tabular}{p{0.10\textwidth} p{0.22\textwidth} p{0.24\textwidth} p{0.20\textwidth} p{0.10\textwidth}}
\toprule
\textbf{Method} & \textbf{Computation Time} & \textbf{Detection Effectiveness} & \textbf{Misclassification Risk} & \textbf{Closed-Weight Compatible} \\
\midrule
Semi-half & Low ($O(n)$) & Low recall and precision & Weak at detecting leaked instances & Yes \\
\midrule
Permutation & Very high ($O(n!)$) & Effective (F1-score $78\%-96\%$) & May misclassify common option questions as leaked & No \\
\midrule
Permutation-R & Medium–high ($O(p \cdot [n!])$) & Competitive with Permutation (F1-score $>80\%$) & Same issue with the common option patterns & No \\
\midrule
Permutation-Q & Moderate ($O(n^2)$) & Effective (F1-score $>82\%$, up to $96\%$ in HellaSwag); often better than original & Same issue with the common option patterns & No \\
\midrule
N-Gram & Depends on token length ($O(m)$ where $m$ is token count) & Very effective (F1-score $>81\%$, up to $100\%$ in HellaSwag) & Same issue with the common option patterns & Yes \\
\bottomrule
\end{tabular}
\caption{Comparison of leakage detection methods across key aspects.\label{tab:pros-cons-restructured}}
\end{table*}

\section{Performance Changes under the Weak Definition of Leakage}
\label{sec:weak-leakage}

\begin{figure}[h]
  \centering
  \includegraphics[width=1\linewidth]{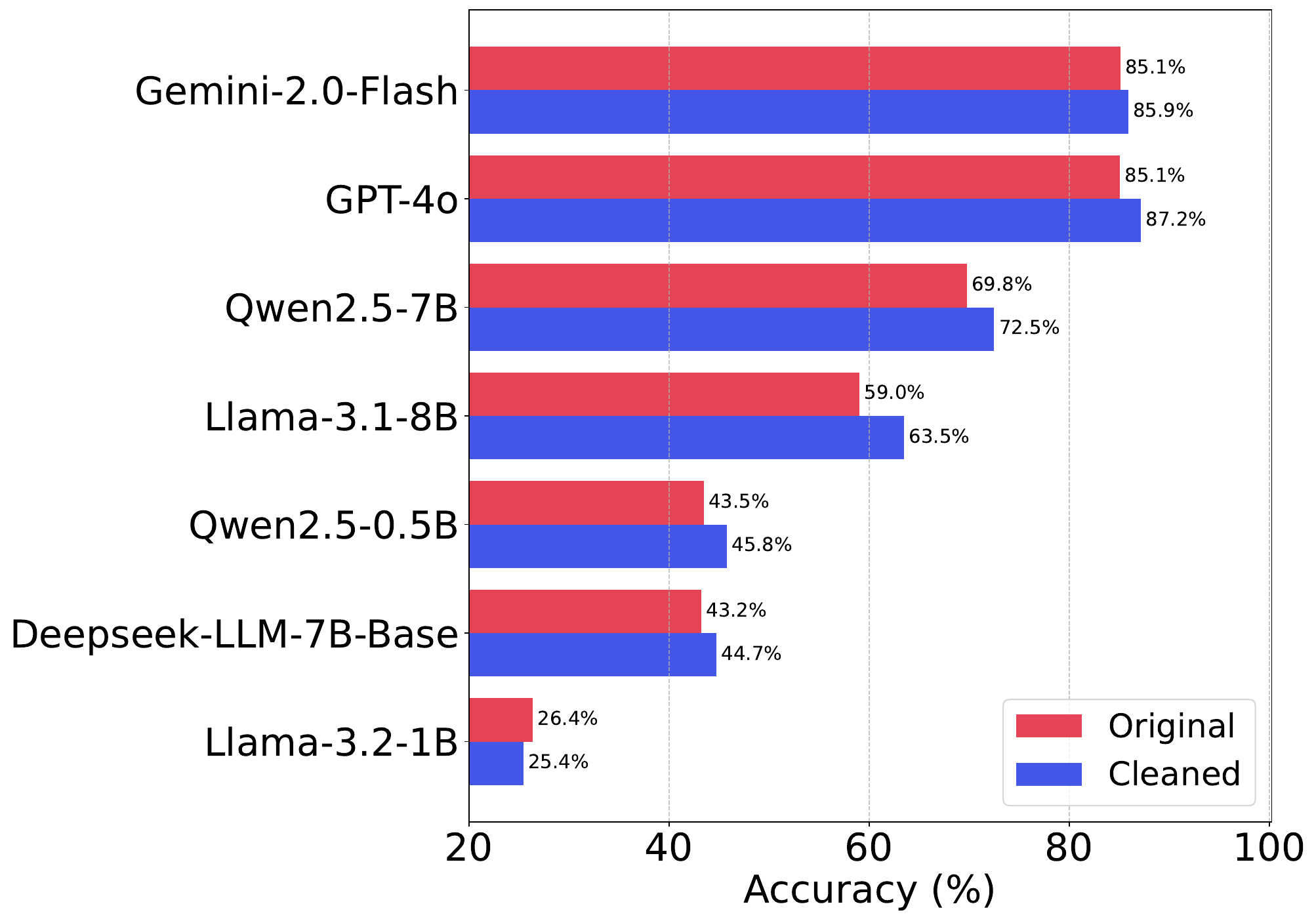}
  \caption{Comparison of model performance on original vs. cleaned MMLU benchmark based on weak definition of leakage.\label{fig:performance-weak}}
\end{figure}

In addition to analyzing performance changes based on the strong definition of leakage, we also examine the shifts that occur under the weak definition. Figure~\ref{fig:performance-weak} presents the performance comparison on the original versus the cleaned dataset under the weak leakage definition.

After cleaning, GPT-4o ranks first, outperforming Gemini-2.0-Flash, while the ranking of the remaining models remains unchanged. Since no model achieves perfect accuracy on leaked instances, removing them leads to a reduction in the proportion of incorrect answers. Consequently, the overall accuracy of all models increases.

\end{document}